# An Internal Cluster Validity Index Using a Distance-based Separability Measure

Shuyue Guan
*Department of Biomedical Engineering*
*The George Washington University*
Washington DC, USA
frankshuyueguan@gwu.edu

Murray Loew*
*Department of Biomedical Engineering*
*The George Washington University*
Washington DC, USA
loew@gwu.edu

*Abstract*—To evaluate clustering results is a significant part of cluster analysis. There are no true class labels for clustering in typical unsupervised learning. Thus, a number of internal evaluations, which use predicted labels and data, have been created. They are also named internal cluster validity indices (CVIs). Without true labels, to design an effective CVI is not simple because it is similar to create a clustering method. And, to have more CVIs is crucial because there is no universal CVI that can be used to measure all datasets, and no specific method for selecting a proper CVI for clusters without true labels. Therefore, to apply more CVIs to evaluate clustering results is necessary. In this paper, we propose a novel CVI – called Distance-based Separability Index (DSI), based on a data separability measure. We applied the DSI and eight other internal CVIs including early studies from Dunn (1974) to most recent studies CVDD (2019) as comparison. We used an external CVI as ground truth for clustering results of five clustering algorithms on 12 real and 97 synthetic datasets. Results show DSI is an effective, unique, and competitive CVI to other compared CVIs. In addition, we summarized the general process to evaluate CVIs and created a new method – rank difference – to compare the results of CVIs.

## I. INTRODUCTION

As a typical unsupervised learning method, cluster analysis plays a critical role in machine learning. Without any prior knowledge, a clustering algorithm could divide a dataset into clusters [1] based on the distribution structure of the data. Clustering is a main study of data mining [2] and widely used in many fields, including pattern recognition [3], object detection [4], image segmentation [5], bioinformatics [6], and data compression [7]. For some machine learning applications, such as medical image analysis, one problem is the shortage of training data because labeling is expensive [8]; the same problem arises in applications of big data [9]. Cluster analysis is a promising solution to the lack of labels problem. Clustering methods can be categorized into centroid-based (*e.g.* k-means), distribution-based (*e.g.* EM algorithm [10]), density-based (*e.g.* DBSCAN [11]), hierarchical (*e.g.* Ward linkage [12]), spectral clustering [13] and others. However, none of the clustering methods can perform well with all datasets [14], [15]. In other words, a clustering method performing well for some types of datasets will perform poorly for some others. Hence, many clustering methods have been created for various types of datasets. As a result, to evaluate which method performs well for a dataset, effective measures of clustering quality (clustering validations) are required [16], [17]. Meanwhile, these measures are also employed to tune parameters of clustering algorithms.

Clustering evaluations have two categories: internal and external evaluations. External evaluations use the true class labels and predicted labels, and internal evaluations use predicted labels and data. Since external evaluations require true labels, they are applied for supervised learning. For absolute unsupervised learning tasks, there are no true class labels, thus we can apply only the internal evaluations for clusters [18]. In fact, to measure clustering results by internal evaluations is as difficult as to analyze clustering itself [19] because measurements have no more information than the clustering methods. Therefore, designing an internal cluster validity index (CVI) is similar to creating an optimizing function for a clustering algorithm. The difference is that the optimizing function will provide a value (loss) with the ability to update clustering results but CVI delivers only a value for reference.

## II. METHODS

### A. Related Work

A variety of CVIs have been designed to deal with the many types of datasets [20]. Here, we select some representative CVIs for comparison. Since external evaluations use the true class labels and predicted labels, we choose an external CVI – the adjusted Rand index (ARI) [21], as the ground truth for comparison.

Internal CVIs can be categorized in two types: center and non-center by their representatives for calculation [22]. Center based CVIs use information of clusters. For example, the Davies–Bouldin index (DB) [23] involves cluster diameters and the distance between cluster centroids. Non-center CVIs use information of data points. For example, the Dunn index [24] uses the minimum and maximum distances between two data points. Besides the DB and Dunn indexes, two more traditional CVIs are applied for comparison in our study: Calinski-Harabasz index (CH) [25] and Silhouette coefficient (Sil) [26]. For recently developed CVIs, we select the I index [27], WB index [28], clustering validation index based on nearest neighbors (CVNN) [18], and cluster validity index based on density-involved distance (CVDD) [22].

In summary, we select one external CVI as the ground truth and eight typical internal CVIs for comparison with our proposed CVI. The compared CVIs range from early studies from Dunn (1974) to the most recent studies CVDD (2019). Unless specifically stated otherwise, **the CVIs that appear in the rest paper mean internal CVIs**. And **the single external CVI is named ARI**.

*B. Distance-based Separability Measure*

Generally, the goal of clustering is to separate a dataset into clusters. From a macro perspective, we consider that the degree of separability of clusters could indicate how well a dataset has been separated. A clustering algorithm assigns every data point a class label. In the worst case, if all labels are assigned randomly, the data points of different classes could be seen having the same distribution. This is the most difficult situation for separation of the dataset. To analyze how the classes of data are distributed, we propose the distance-based separability index (DSI).

If two classes $X$ and $Y$ have have $N_x, N_y$ points, we define that the **Intra-Class distance (ICD) set** is the set of distances between any two points in the same class $X$:

$$\{d_x\} = \left\{\|x_i - x_j\|_2 \mid x_i, x_j \in X; x_i \neq x_j\right\}$$

If $|X| = N_x$, then $|\{d_x\}| = \frac{1}{2}N_x(N_x - 1)$.

And the **Between-class distance (BCD) set** is the set of distances between any two points from different classes $X$ and $Y$:

$$\{d_{x,y}\} = \left\{\|x_i - y_j\|_2 \mid x_i \in X; x_j \in Y\right\}$$

If $|X| = N_x, |Y| = N_y$, then $|\{d_{x,y}\}| = N_x N_y$.

It can be shown[1] that: if and only if two classes $X$ and $Y$ have sufficient data points with the same distribution, the distributions of ICD and BCD sets are nearly identical. Hence, if the distributions of the ICD and BCD sets are nearly identical, all labels are assigned randomly and thus the dataset has the worst separability. The metric of distance is Euclidean ($l^2$-norm). The time cost for computing ICD and BCD sets is linear with the dimensionality, and quadratic with the number, of observations.

To compute the DSI of the two classes $X$ and $Y$, first, we compute the ICD sets of $X$ and $Y$: $\{d_x\}, \{d_y\}$ and the BCD set: $\{d_{x,y}\}$. To examine the similarity of the distributions of ICD and BCD sets, we apply the Kolmogorov–Smirnov (KS) test [29]. Although there are many statistical measures to compare two distributions, such as Bhattacharyya distance, Kullback–Leibler divergence and Jensen–Shannon divergence, most require that the two sets have the same number of observations. It is easy to show that the $|\{d_x\}|, |\{d_y\}|$ and $|\{d_{x,y}\}|$ cannot be the same. The similarities between the ICD and BCD sets are then computed using the KS test: $s_x = KS(\{d_x\}, \{d_{x,y}\})$ and $s_y = KS(\{d_y\}, \{d_{x,y}\})$. Since there are two classes, the DSI is the average of the two KS similarities: $DSI(\{X, Y\}) = (s_x + s_y)/2$. The $KS(\{d_x\}, \{d_y\})$ is not considered because it shows only the difference of distribution shapes, not their location information. For example, two distributions that have the same shape, but no overlap will have similarities between ICD: $KS(\{d_x\}, \{d_y\})$ equal to zero. In general, for an *n*-class dataset (**DSI Algorithm**):

- Compute *n* ICD sets for each class: $\{d_{C_i}\}; i = 1, 2, \cdots, n$.

- For the *i*-th class of data $C_i$, the BCD set is the set of distances between any two points in $C_i$ and $\bar{C}_i$ (other classes, not $C_i$): $\{d_{C_i, \bar{C}_i}\}$. The KS similarity between ICD and BCD set is $s_i = KS(\{d_{C_i}\}, \{d_{C_i, \bar{C}_i}\})$.

- The DSI of this dataset is $DSI(\{C_i\}) = (\sum s_i)/n$.

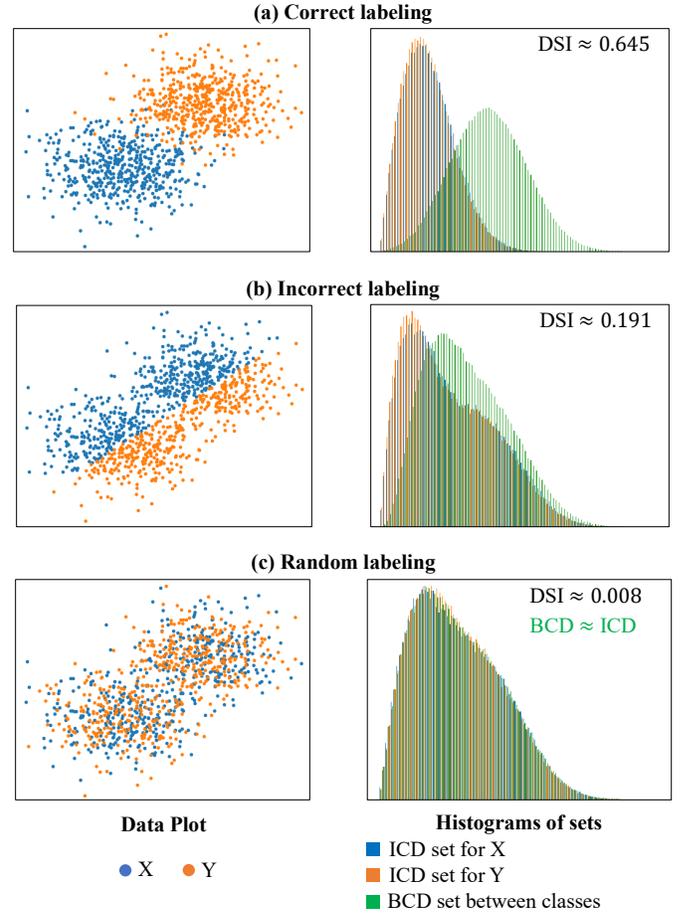

Figure 1. Two clusters (classes) datasets with different label assignments. Each histogram indicates the relative frequency of the value of each of the three distance measures (indicated by color).

---

[1] We do not show the proof here because it is detailed and not relevant to the main topic. It will appear in another forthcoming publication, which can be found in author's website linked up with the ORCID: https://orcid.org/0000-0002-3779-9368. Since the statement is intuitive, we provide an informal explanation here: points in $X$ and $Y$ having the same distribution and covering the same region can be considered to have been sampled from one distribution $Z$. Hence, both ICDs of $X$ and $Y$, and BCDs between $X$ and $Y$ are actually ICDs of $Z$. Consequently, the distributions of ICDs and BCDs are identical.

Figure 1 displays a two-class dataset. If the labels are assigned correctly, the distributions of ICD sets will be different from the BCD set; the DSI will reach the maximum value for this dataset (see Figure 1a) because the two clusters are well separated. For incorrect clustering (Figure 1b), the difference between distributions of ICD and BCD sets becomes smaller so that the DSI value decreases. For an extreme situation (Figure 1c), if all labels are randomly assigned, the distributions of the ICD and BCD sets will be nearly the same. This is the worst case of separation; DSI is close to zero. Therefore, the DSI could well reflect the separability of clusters. DSI ranges from 0 to 1; the greater value means the dataset is clustered better.

## III. EXPERIMENTS

The goal of CVIs is to evaluate the clustering results. In this study, we employ several internal CVIs including our proposed DSI to examine the clustering results from several clustering methods (algorithms). For a given dataset, different clustering algorithms may return different clusters; CVIs are used to find the best clusters. We choose an external CVI – the adjusted Rand index (ARI) as the ground truth for comparison because ARI involves true labels (clusters) of the dataset. Eight commonly used CVIs are selected to compare with the proposed DSI and are shown in TABLE I. They include classical and recent CVIs.

TABLE I    COMPARED CVIs.

| Name | Optimal[a] | Reference |
|---|---|---|
| **D**unn index | Max | (Dunn, J.,1973) [24] |
| **C**alinski-**H**arabasz Index | Max | (Calinski & Harabasz, 1974) [25] |
| **D**avies–**B**ouldin index | min | (Davies & Bouldin, 1979) [23] |
| **Sil**houette Coefficient | Max | (Rousseeuw, 1987) [26] |
| **I** | Max | (U. Maulik, 2002) [27] |
| **WB** | min | (Zhao Q., 2009) [28] |
| **CVNN** | min | (Yanchi L., 2013) [18] |
| **CVDD** | Max | (Lianyu H., 2019) [22] |
| **DSI** | Max | *Proposed* |

[a.] Optimal column means the CVI for best case has the minimum or maximum value.

### A. Synthetic datasets

The synthetic datasets for clustering are from the Tomas Barton repository. This repository contains 122 artificial datasets[2]. Each dataset has hundreds to thousands of objects with several to tens of classes in two or three dimensions (features). We selected 97 datasets for experiment; the 25 unused datasets have too many objects to run the clustering processing in reasonable time. The names of the 97 used synthetic datasets are shown in TABLE II and illustrated in Tomas Barton's homepage[3].

TABLE II    NAMES OF THE USED SYNTHETIC DATASETS.

| | | | | |
|---|---|---|---|---|
| 3-spiral | 2d-10c | ds2c2sc13 | rings | square5 |
| aggregation | 2d-20c-no0 | ds3c3sc6 | shapes | st900 |
| 2d-3c-no123 | threenorm | ds4c2sc8 | simplex | target |
| dense-disk-3000 | triangle1 | 2d-4c | sizes1 | tetra |
| dense-disk-5000 | triangle2 | 2dnormals | sizes2 | curves1 |
| elliptical_10_2 | dartboard1 | engytime | sizes3 | curves2 |
| elly-2d10c13s | dartboard2 | flame | sizes4 | D31 |
| 2sp2glob | 2d-4c-no4 | fourty | sizes5 | twenty |
| cure-t0-2000n-2D | 2d-4c-no9 | gaussians1 | smile1 | aml28 |
| cure-t1-2000n-2D | diamond9 | hepta | smile2 | wingnut |
| twodiamonds | disk-1000n | hypercube | smile3 | xclara |
| spherical_4_3 | disk-3000n | jain | atom | xor |
| spherical_5_2 | disk-4000n | long1 | blobs | zelnik1 |
| spherical_6_2 | disk-4500n | long2 | cassini | zelnik2 |
| chainlink | disk-4600n | long3 | spiral | zelnik3 |
| spiralsquare | disk-5000n | longsquare | circle | zelnik5 |
| complex8 | donut1 | lsun | square1 | zelnik6 |
| complex9 | donut2 | pathbased | square2 | |
| compound | donut3 | pmf | square3 | |
| donutcurves | cuboids | R15 | square4 | |

### B. Real datasets

In this study, the 12 real datasets used for clustering are from three sources: the *sklearn.datasets* package[4], UC Irvine Machine Learning Repository [30] and Tomas Barton's repository (real world datasets). Unlike the synthetic datasets, the dimensions (feature numbers) of most selected real datasets are greater than three. Hence, to evaluate their clustering results we must use CVIs rather than plotting clusters as for 2D or 3D synthetic datasets. Details about the 12 real datasets appear in TABLE III.

TABLE III    THE DESCRIPTION OF USED REAL DATASETS.

| Name | Title | Object# | Feature# | Class# |
|---|---|---|---|---|
| Iris | Iris plants dataset | 150 | 4 | 3 |
| digits | Optical recognition of handwritten digits dataset | 5620 | 64 | 10 |
| wine | Wine recognition dataset | 178 | 13 | 3 |
| cancer | Breast cancer Wisconsin (diagnostic) dataset | 569 | 30 | 2 |
| faces | Olivetti faces data-set | 400 | 4096 | 40 |
| vertebral | Vertebral Column Data Set | 310 | 6 | 3 |
| haberman | Haberman's Survival Data | 306 | 3 | 2 |
| sonar | Sonar, Mines vs. Rocks | 208 | 60 | 2 |
| tae | Teaching Assistant Evaluation | 151 | 5 | 3 |
| thy | Thyroid Disease Data Set | 215 | 5 | 3 |
| vehicle | Vehicle silhouettes | 946 | 18 | 4 |
| zoo | Zoo Data Set | 101 | 16 | 7 |

---

[2] https://github.com/deric/clustering-benchmark/tree/master/src/main/resources/datasets/artificial
[3] https://github.com/deric/clustering-benchmark
[4] https://scikit-learn.org/stable/datasets/index.html#

## C. Results and Evaluations

In summary, the steps to verify CVIs are:

- To run different **clustering algorithms** on a dataset to obtain different clustering results.
- To compute **CVIs** of these clustering results and their ARI with real labels.
- Then **to compare** the values of CVI with **ARI** (ground truth) [21].
- Repeat the former three steps for a new **dataset**.

Five clustering algorithms from various categories are used, they are: k-means, Ward linkage, spectral clustering, BIRCH [31] and EM algorithm (Gaussian Mixture). CVIs used for verification and comparison are shown in TABLE I and the datasets used are shown in TABLE II and TABLE III.

The main problem is how to compare the values of CVIs with ground truth ARI. There are two plans:

*1) Hit-the-best.* Clustering results of different clustering algorithms on a dataset would have different CVIs and ARI; if a CVI gives the best score to a clustering result that also has the best ARI score, this CVI is considered to be hit-the-best (correct prediction).

TABLE IV shows an example of hit-the-best. For the "wine" dataset, k-means receives the best ARI score and Dunn, DB, WB, I, CVNN and DSI give k-means the best score; and thus, the six CVIs are hit-the-best.

TABLE IV      CLUSTERING RESULTS AND THEIR CVI SCORES ON THE WINE RECOGNITION DATASET.

| Clustering method<br>Validity[a] | KMeans | Ward Linkage | Spectral Clustering | BIRCH | EM |
|---|---|---|---|---|---|
| ARI[b] + | **0.913**[c] | 0.757 | 0.880 | 0.790 | 0.897 |
| Dunn + | **0.232** | 0.220 | 0.177 | 0.229 | **0.232** |
| CH + | 70.885 | 68.346 | 70.041 | 67.647 | **70.940** |
| DB - | **1.388** | 1.390 | 1.391 | 1.419 | 1.389 |
| Silhouette + | 0.284 | 0.275 | 0.283 | 0.277 | **0.285** |
| WB - | **3.700** | 3.841 | 3.748 | 3.880 | **3.700** |
| I + | **5.421** | 4.933 | 5.326 | 4.962 | **5.421** |
| CVNN - | **21.859** | 22.134 | 21.932 | 22.186 | **21.859** |
| CVDD + | 31.114 | **31.141** | 29.994 | 30.492 | 31.114 |
| DSI + | **0.635** | 0.606 | 0.629 | 0.609 | 0.634 |

a. CVI for best case has the minimum (-) or maximum (+) value.
b. Gray background is the only row of ARI as ground truth; other rows are CVIs.
c. **Bold value**: the best case by the measure of this row.

For the hit-the-best plan, however, the best score can be relatively unstable and random in some cases. For example, in TABLE IV, the ARI score of EM is very close to that of k-meams and the Silhouette score of EM is also very close to that of k-means. If these values fluctuated a little and changed the best cases, the comparison outcome for this dataset will be changed. Another drawback of this plan is that it only concerns one best case and ignores others. A hit-the-best plan does not evaluate the whole picture for one dataset. It might be a more strict criterion but lacks robustness. It is vulnerable to extreme cases such as when scores of different clustering results are very close to each other. Hence, we create another plan to compare the score sequences of CVIs and ARI through their orders.

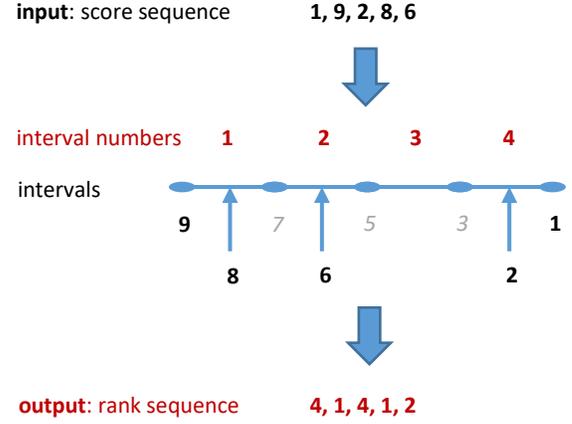

Figure 2. An example of rank numbers assignment.

*2) Rank difference.* This comparison plan fixes the two problems in the hit-the-best plan. One is instablility for similar scores and the other one is the bias on only one case.

We apply quantization to solve the problem of similar scores. Every score in the score sequence of a CVI (*i.e.,* a row in TABLE IV) will be assigned a rank number and similar scores have high probability to be allocated the same rank number. The procedure is:

- Find the minimum and maximum values of $N$ scores from one sequence.
- Uniformly divide $[min, max]$ into $N-1$ intervals.
- Label intervals from $max$ to $min$ by $1, 2, \ldots, N-1$.
- If a score is in the $k$-th interval, its rank number is $k$.
- Define rank number of $max$ is 1, and intervals are left open and right closed: (left, right].

Figure 2 shows an example of converting a score sequence to rank numbers (a rank sequence). The rank number of scores 9 and 8 is 1 because they are in the 1st interval. For the same reason, the rank number of scores 1 and 2 is 4. Such quantization is better than assigning rank numbers by ordering because it avoids the assignment of different rank numbers to very close scores in most cases (it is still possible to use different rank numbers for very close scores; for example, in the Figure 2 case, if scores 8 and 6 changed to 7.1 and 6.9, their rank numbers will still be 1 and 2 even they are very close).

For two score sequences (*e.g.* CVI and ARI), after quantizing them to two rank sequences, we will compute the difference of two rank sequences (called the rank difference), which is simply defined as the summation of absolute difference between two rank sequences. For example, if two rank sequences are:

$R1: \{4, 1, 4, 1, 2\}; R2: \{1, 3, 2, 4, 4\};$

Summation of absolute difference is:

$|4 - 1| + |1 - 3| + |4 - 2| + |1 - 4| + |2 - 4| = 12$

Smaller rank difference means the distance of two sequences is closer. That two sequences of CVI and ARI are closer indicates a better prediction. It is not difficult to show that rank difference for two N-length score sequences ranges: $[0, N(N - 2)]$.

## IV. RESULTS

As discussed before, for one dataset and a CVI, an evaluation result can be computed by using the hit-the-best or rank difference comparison. In other words, one result is obtained by comparing one CVI row in TABLE IV with the gray-background row (ARI). The outcome of hit-the-best comparison is either 0 or 1. The 1 means that the best clusters predicted by CVI are the same as ARI; otherwise the outcome is 0. The outcome of the rank difference comparison is a value in the range $[0, N(N - 2)]$. As TABLE IV shows, the length of score sequences is 5; hence, the range of rank difference is $[0, 15]$. The smaller value means the CVI predicts better.

We applied this evaluation to all datasets (real and synthetic) and selected CVIs. TABLE V and TABLE VII are hit-the-best comparison results for real and synthetic datasets[5]. TABLE VI and TABLE VIII are rank difference comparison results for real and synthetic datasets. To compare across data sets, we summed all results in the last row. For the hit-the-best comparison, the larger total value is better because more hits appear. For the rank difference comparison, the smaller total value is better because results of the CVI are closer to that of ARI. Finally, ranks in the last row uniformly indicate CVIs' performances. The smaller rank number means better performance.

TABLE V    HIT-THE-BEST RESULTS FOR REAL DATASETS.

| Dataset | Dunn | CH | DB | Silhouette | WB | I | CVNN | CVDD | DSI |
|---|---|---|---|---|---|---|---|---|---|
| Iris | 0 | 0 | 0 | 0 | 0 | 0 | 0 | 1 | 0 |
| digits | 0 | 0 | 0 | 1 | 0 | 0 | 1 | 0 | 1 |
| wine | 1 | 0 | 1 | 0 | 1 | 1 | 1 | 0 | 1 |
| cancer | 0 | 0 | 0 | 0 | 0 | 0 | 1 | 0 | 0 |
| faces | 1 | 1 | 1 | 1 | 1 | 1 | 0 | 1 | 1 |
| vertebral | 0 | 0 | 0 | 0 | 0 | 0 | 0 | 0 | 0 |
| haberman | 0 | 1 | 0 | 0 | 1 | 0 | 0 | 0 | 0 |
| sonar | 0 | 1 | 0 | 0 | 1 | 0 | 0 | 0 | 0 |
| tae | 0 | 0 | 0 | 0 | 0 | 0 | 1 | 1 | 0 |
| thy | 0 | 0 | 0 | 0 | 0 | 0 | 0 | 0 | 0 |
| vehicle | 0 | 0 | 0 | 0 | 0 | 0 | 1 | 0 | 1 |
| zoo | 1 | 0 | 1 | 0 | 0 | 1 | 0 | 0 | 1 |
| Total[a] (rank) | 3 (4) | 3 (4) | 3 (4) | 2 (9) | 4 (3) | 3 (4) | 5 (1) | 3 (4) | 5 (1) |

a. Larger value is better (smaller rank).

TABLE VI    RANK DIFFERENCE RESULTS FOR REAL DATASETS.

| Dataset | Dunn | CH | DB | Silhouette | WB | I | CVNN | CVDD | DSI |
|---|---|---|---|---|---|---|---|---|---|
| Iris | 8 | 13 | 15 | 15 | 13 | 11 | 15 | 6 | 15 |
| digits | 2 | 2 | 1 | 1 | 4 | 6 | 8 | 7 | 6 |
| wine | 9 | 1 | 3 | 1 | 1 | 0 | 0 | 7 | 0 |
| cancer | 8 | 7 | 6 | 9 | 7 | 8 | 2 | 7 | 9 |
| faces | 4 | 3 | 4 | 4 | 2 | 3 | 9 | 2 | 5 |
| vertebral | 6 | 13 | 14 | 12 | 15 | 13 | 15 | 6 | 13 |
| haberman | 9 | 7 | 7 | 7 | 7 | 9 | 7 | 7 | 8 |
| sonar | 7 | 3 | 3 | 4 | 3 | 4 | 11 | 10 | 3 |
| tae | 9 | 14 | 9 | 9 | 14 | 15 | 0 | 9 | 9 |
| thy | 5 | 2 | 2 | 2 | 2 | 6 | 2 | 3 | 10 |
| vehicle | 12 | 11 | 9 | 13 | 13 | 12 | 3 | 3 | 7 |
| zoo | 1 | 6 | 1 | 6 | 6 | 1 | 9 | 8 | 1 |
| Total[a] (rank) | 80 (3) | 82 (5) | 74 (1) | 83 (6) | 87 (8) | 88 (9) | 81 (4) | 75 (2) | 86 (7) |

a. Smaller value is better (smaller rank).

---

[5] The code can be found in author's website linked up with the ORCID: https://orcid.org/0000-0002-3779-9368

TABLE VII   HIT-THE-BEST RESULTS FOR 97 SYNTHETIC DATASETS.

| Dataset | Dunn | CH | DB | Silhouette | WB | I | CVNN | CVDD | DSI |
|---|---|---|---|---|---|---|---|---|---|
| 3-spiral | 1 | 0 | 0 | 0 | 0 | 0 | 0 | 1 | 0 |
| aggregation | 0 | 0 | 0 | 0 | 0 | 0 | 1 | 1 | 1 |
| … | … | … | … | … | … | … | … | … | … |
| zelnik5 | 1 | 0 | 0 | 0 | 0 | 0 | 0 | 1 | 0 |
| zelnik6 | 1 | 1 | 0 | 0 | 1 | 0 | 0 | 0 | 0 |
| Total[a] (rank) | 46 (2) | 30 (8) | 35 (4) | 35 (4) | 29 (9) | 31 (7) | 35 (4) | 50 (1) | 40 (3) |

[a.] Larger value is better (smaller rank).

TABLE VIII   RANK DIFFERENCE RESULTS FOR 97 SYNTHETIC DATASETS.

| Dataset | Dunn | CH | DB | Silhouette | WB | I | CVNN | CVDD | DSI |
|---|---|---|---|---|---|---|---|---|---|
| 3-spiral | 2 | 12 | 14 | 13 | 14 | 12 | 13 | 1 | 13 |
| aggregation | 3 | 3 | 2 | 2 | 4 | 5 | 2 | 5 | 3 |
| … | … | … | … | … | … | … | … | … | … |
| zelnik5 | 4 | 10 | 12 | 10 | 11 | 11 | 10 | 4 | 11 |
| zelnik6 | 4 | 3 | 2 | 2 | 3 | 3 | 5 | 2 | 2 |
| Total[a] (rank) | 406 (2) | 541 (6) | 547 (7) | 489 (4) | 583 (9) | 554 (8) | 504 (5) | 337 (1) | 415 (3) |

[a.] Smaller value is better (smaller rank).

## V. DISCUSSION

Although DSI obtains only one top-1 rank (TABLE V) in these experiments, having no last rank means it performed better than some other CVIs. It is worth emphasizing that all compared CVIs are excellent and widely used. Therefore, experiments show that DSI can join them as a new promising CVI. Actually, by examining those CVI evaluation results, we confirm that no one CVI performs well for all datasets. And thus, it is better to measure clustering results by using more effective CVIs. The DSI provides another CVI option. Also, DSI is unique: no other CVIs perform the same as DSI. For example, in TABLE V, for the "vehicle" dataset, only CVNN and DSI predicted correctly. But for "zoo" dataset, CVNN was wrong and DSI was correct. For another example in TABLE VI, for the "sonar" dataset, DSI performed better than Dunn, CVNN, and CVDD; but for the "cancer" dataset, Dunn, CVNN, and CVDD performed better than DSI. More examples of the diversity of CVI are shown in TABLE IX and their plots with true labels are shown in Figure 3 (the "atom" dataset has three features; the others have two features).

The former examples show the need for employing more CVIs because each is different and every CVI may have its special capability. That capability, however, is difficult to describe clearly. Some CVIs' definitions show them to be categorized into center/non-center representative [22] or density-representative. Similarly, the DSI is a separability-representative CVI. That is, DSI performs better for clusters having high separability with true labels (like the "atom" dataset in in Figure 3); otherwise, if real clusters have low separability, the incorrectly predicted clusters may have a higher DSI score (Figure 4).

TABLE IX   RANK DIFFERENCE RESULTS FOR SELECTED SYNTHETIC DATASETS.

| Dataset | Dunn | CH | DB | Sil | WB | I | CVNN | CVDD | DSI |
|---|---|---|---|---|---|---|---|---|---|
| atom | 0 | 15 | 15 | 15 | 15 | 14 | 4 | 0 | 0 |
| disk-4000n | 10 | 0 | 7 | 0 | 0 | 0 | 11 | 12 | 1 |
| disk-1000n | 6 | 12 | 15 | 12 | 13 | 14 | 15 | 8 | 14 |
| D31 | 5 | 1 | 2 | 1 | 0 | 2 | 10 | 2 | 0 |
| flame | 10 | 6 | 11 | 7 | 7 | 8 | 12 | 11 | 7 |
| square3 | 11 | 0 | 2 | 0 | 0 | 7 | 0 | 11 | 0 |

Clusters in datasets have great diversity so that the diversity of clustering methods and CVIs is necessary. Since the preferences of CVIs are difficult to analyze precisely and quantitatively, more studies for selecting a proper CVI to measure clusters without true labels need to be done in the future. More CVIs expand the options. And, before that breakthrough, it is meaningful to provide more effective CVIs and apply more than one CVIs to evaluate clustering results.

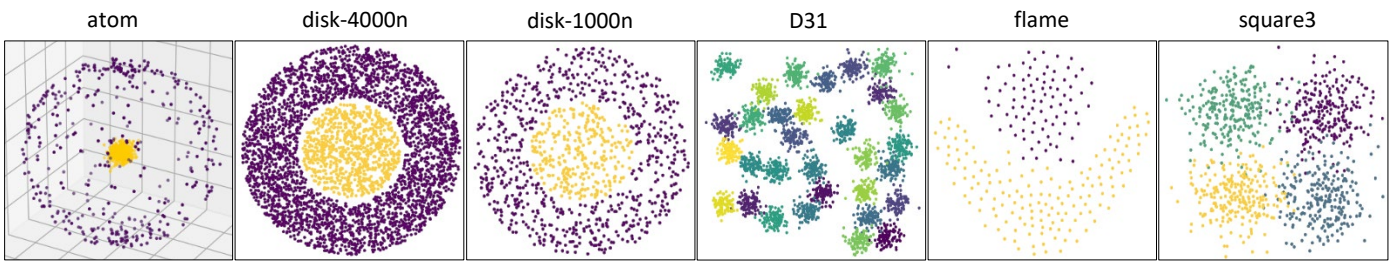

Figure 3. Examples for rank differences of synthetic datasets.

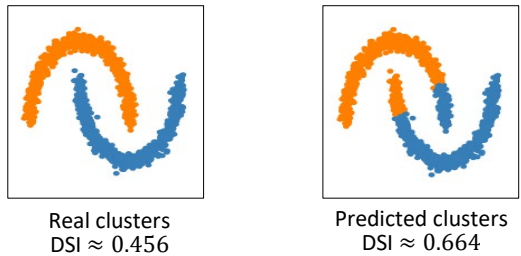

Figure 4. Wrongly predicted clusters have a higher DSI score than real clusters.

In addition, to evaluate CVIs is also an important work. The general process is to:

- Create different clusters from datasets;
- Compute external CVI with true labels as ground truth and internal CVIs;
- Compare results of internal CVIs with the ground truth. Results from an effective internal CVI should be close to the results of an external CVI.

In this paper, we generated different clusters using different clustering methods. This step can also be achieved through changing parameters of clustering algorithms (like the $k$ in k-means clustering) or taking subsets of datasets. The comparison step also has alternative methods. For example, besides the two plans we used, to compare the optimal number of clusters recognized by CVIs [32] is another feasible plan.

## VI. CONCLUSION

Since there is no universal CVI for all datasets and no specific method for selecting a proper CVI to measure clusters without true labels, to apply more CVIs to evaluate clustering results is inevitable. In this paper, we propose a novel CVI, called DSI, based on a data separability measure. The goal of clustering is to separate a dataset into clusters; and we hypothesize that better clustering could cause these clusters to have a higher separability.

We applied nine internal CVIs including the proposed DSI, and an external CVI (ARI) as ground truth to clustering results of five clustering algorithms on various datasets. The results show DSI is an effective, unique, and competitive CVI to other compared CVIs. We summarized the general process to evaluate CVIs and used two methods to compare the results of CVIs with ground truth scores. The second comparison method – rank difference – which we created, avoids two disadvantages of the hit-the-best method, which is commonly used in CVI evaluation. Both DSI and the new comparison method can be helpful in clustering analysis and CVI studies in the future.